\documentclass[preprint,12pt]{elsarticle}

\usepackage{listings}
\usepackage{color}

\definecolor{dkgreen}{rgb}{0,0.6,0}
\definecolor{gray}{rgb}{0.5,0.5,0.5}
\definecolor{mauve}{rgb}{0.58,0,0.82}

\usepackage{tikz}
\usetikzlibrary{backgrounds}
\usepackage{graphicx}
\usepackage{bm}
\usepackage{amsmath}
\usepackage{setspace}
\usepackage{hyperref}
\usepackage{subfigure}
\usepackage{multirow}
\hypersetup{
    colorlinks=true,       
}
\usepackage{amssymb}

\usepackage{lineno}
\usepackage{diagbox}

\journal{Journal Name}

\begin{document}

\begin{frontmatter}


\title{Physics Informed Deep Learning (Part II): Data-driven Discovery of Nonlinear Partial Differential Equations}



\author{Maziar Raissi$^{1}$, Paris Perdikaris$^{2}$, and George Em Karniadakis$^{1}$}
\address{$^{1}$Division of Applied Mathematics, Brown University,\\ Providence, RI, 02912, USA\\
$^{2}$Department of Mechanical Engineering and Applied Mechanics,\\ University of Pennsylvania,\\ Philadelphia, PA, 19104, USA}

\begin{abstract}

We introduce {\em physics informed neural networks} -- neural networks that are trained to solve supervised learning tasks while respecting any given law of physics described by general nonlinear partial differential equations. In this second part of our two-part treatise, we focus on the problem of data-driven discovery of partial differential equations. Depending on whether the available data is scattered in space-time or arranged in fixed temporal snapshots, we introduce two main classes of algorithms, namely continuous time and discrete time models. The effectiveness of our approach is demonstrated using a wide range of benchmark problems in mathematical physics, including conservation laws, incompressible fluid flow, and the propagation of nonlinear shallow-water waves.

\end{abstract}


\end{frontmatter}


\section{Introduction}

Deep learning has gained unprecedented attention over the last few years, and deservedly so, as it has introduced transformative solutions across diverse scientific disciplines \cite{krizhevsky2012imagenet,lecun2015deep,lake2015human,alipanahi2015predicting}. Despite the ongoing success, there exist many scientific applications that have yet failed to benefit from this emerging technology, primarily due to the high cost of data acquisition. It is well known that the current state-of-the-art machine learning tools  (e.g., deep/convolutional/recurrent neural networks) are lacking robustness and fail to provide any guarantees of convergence when operating in the {\em small data} regime, i.e., the regime where very few training examples are available. \\

In the first part of this study, we introduced {\em physics informed neural networks} as a viable solution for training deep neural networks with few training examples, for cases where the available data is known to respect a given physical law described by a system of partial differential equations. Such cases are abundant in the study of physical, biological, and engineering systems, where longstanding developments of mathematical physics have shed tremendous insight on how such systems are structured, interact, and dynamically evolve in time. We saw how the knowledge of an underlying physical law can introduce structure that effectively regularizes the training of neural networks, and enables them to generalize well even when only a few training examples are available. Through the lens of different benchmark problems, we highlighted the key features of {\em physics informed neural networks} in the context of data-driven solutions of partial differential equations \cite{raissi2017numerical, raissi2017inferring}.\\

In this second part of our study, we shift our attention to the problem of data-driven discovery of partial differential equations \cite{raissi2017hidden,raissi2017machine, Rudye1602614}. To this end, let us consider parametrized and nonlinear partial differential equations of the general form
\begin{eqnarray}\label{eq:PDE}
&&u_t + \mathcal{N}[u;\lambda] = 0,\ x \in \Omega, \ t\in[0,T],
\end{eqnarray}
where $u(t,x)$ denotes the latent (hidden) solution, $\mathcal{N}[\cdot;\lambda]$ is a nonlinear operator parametrized by $\lambda$, and $\Omega$ is a subset of $\mathbb{R}^D$. This setup encapsulates a wide range of problems in mathematical physics including conservation laws, diffusion processes, advection-diffusion-reaction systems, and kinetic equations. As a motivating example, the one dimensional Burgers' equation \cite{basdevant1986spectral} corresponds to the case where $\mathcal{N}[u;\lambda] = \lambda_1 u u_x - \lambda_2 u_{xx}$ and $\lambda = (\lambda_1, \lambda_2)$. Here, the subscripts denote partial differentiation in either time or space. Now, the problem of data-driven discovery of partial differential equations poses the following question: given a small set of scattered and potentially noisy observations of the hidden state $u(t,x)$ of a system, what are the parameters $\lambda$ that best describe the observed data?\\

In what follows, we will provide an overview of our two main approaches to tackle this problem, namely continuous time and discrete time models, as well as a series of results and systematic studies for a diverse collection of benchmarks. In the first approach, we will assume availability of scattered and potential noisy measurements across the entire spatio-temporal domain. In the latter, we will try to infer the unknown parameters $\lambda$ from only two data snapshots taken at distinct time instants. All data and codes used in this manuscript are publicly available on GitHub at \url{https://github.com/maziarraissi/PINNs}.

\section{Continuous Time Models}
We define $f(t,x)$ to be given by the left-hand-side of equation \eqref{eq:PDE}; i.e.,
\begin{equation}
f := u_t + \mathcal{N}[u;\lambda],\label{eq:PDE_RHS}
\end{equation}
and proceed by approximating $u(t,x)$ by a deep neural network. This assumption along with equation (\ref{eq:PDE_RHS}) result in a \emph{physics informed neural network} $f(t,x)$. This network can be derived by applying the chain rule for differentiating compositions of functions using automatic differentiation \cite{baydin2015automatic}. It is worth highlighting that the parameters of the differential operator $\lambda$ turn into parameters of the \emph{physics informed neural network} $f(t,x)$.

\subsection{Example (Burgers' Equation)}
As a first example, let us consider the Burgers' equation. This equation arises in various areas of applied mathematics, including fluid mechanics, nonlinear acoustics, gas dynamics, and traffic flow \cite{basdevant1986spectral}. It is a fundamental partial differential equation and can be derived from the Navier-Stokes equations for the velocity field by dropping the pressure gradient term. For small values of the viscosity parameters, Burgers' equation can lead to shock formation that is notoriously hard to resolve by classical numerical methods. In one space dimension the equation reads as
\begin{eqnarray}\label{eq:Burgers}
&& u_t + \lambda_1 u u_x - \lambda_2 u_{xx} = 0.
\end{eqnarray}
Let us define $f(t,x)$ to be given by
\begin{eqnarray}\label{eq:Burgers_RHS}
f := u_t + \lambda_1 u u_x - \lambda_2 u_{xx},
\end{eqnarray}
and proceed by approximating $u(t,x)$ by a deep neural network. This will result in the \emph{physics informed neural network} $f(t,x)$. The shared parameters of the neural networks $u(t,x)$ and $f(t,x)$ along with the parameters $\lambda = (\lambda_1, \lambda_2)$ of the differential operator can be learned by minimizing the mean squared error loss
\begin{equation}\label{eq:MSE_Burgers_CT_inference}
MSE = MSE_u + MSE_f,
\end{equation}
where
\[
MSE_u = \frac{1}{N}\sum_{i=1}^{N} |u(t^i_u,x_u^i) - u^i|^2,
\]
and
\[
MSE_f = \frac{1}{N}\sum_{i=1}^{N}|f(t_u^i,x_u^i)|^2.
\]
Here, $\{t_u^i, x_u^i, u^i\}_{i=1}^{N}$ denote the training data on $u(t,x)$. The loss $MSE_u$ corresponds to the training data on $u(t,x)$ while $MSE_f$ enforces the structure imposed by equation \eqref{eq:Burgers} at a finite set of collocation points, whose number and location is taken to be the same as the training data.\\

\begin{figure}[!t]
\includegraphics[width = 1.0\textwidth]{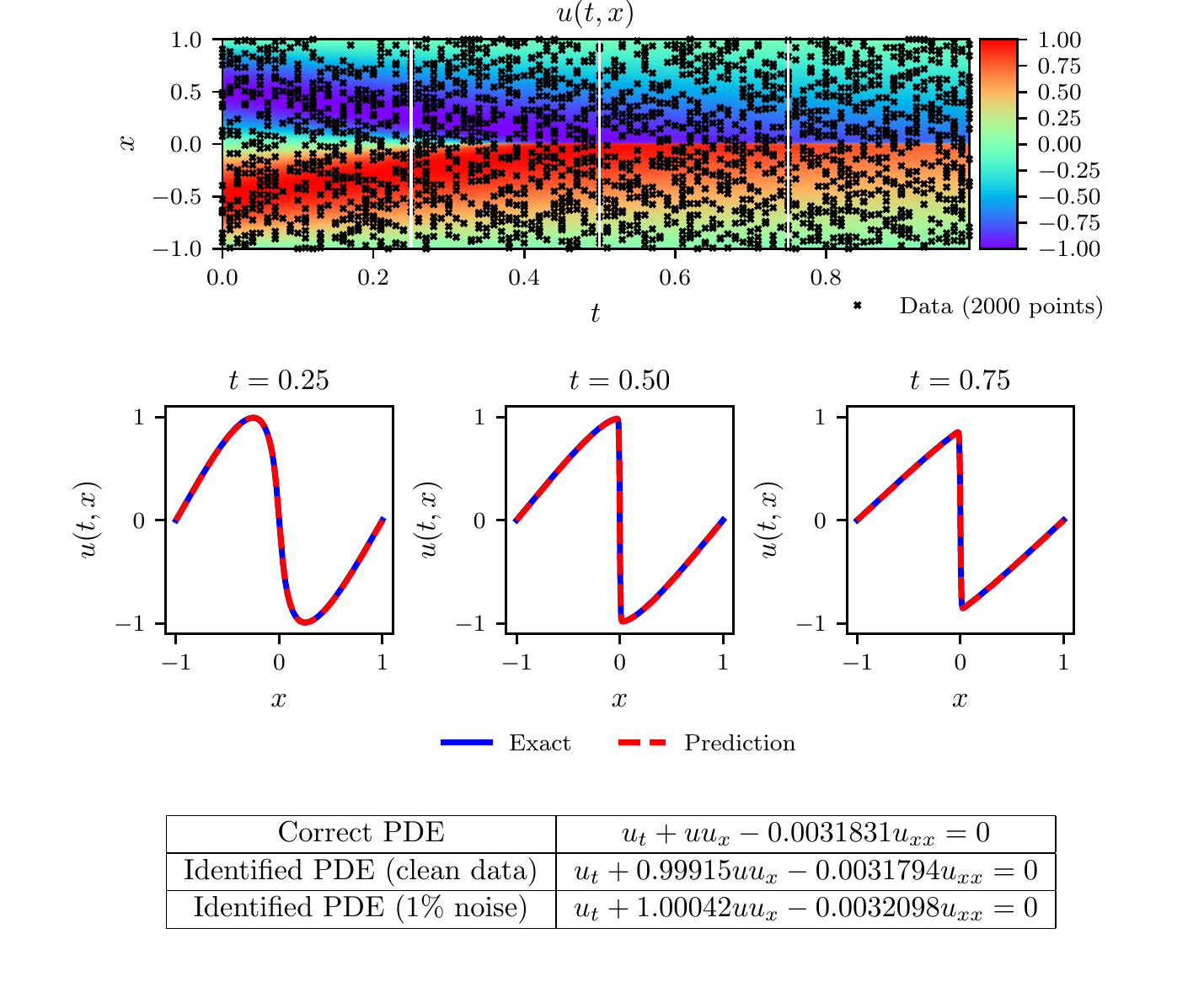}
\vspace*{-1.0cm}
\caption{{\em Burgers equation:} {\it Top:} Predicted solution $u(t,x)$ along with the training data. {\it Middle:} Comparison of the predicted and exact solutions corresponding to  the three temporal snapshots depicted by the dashed vertical lines in the top panel. {\it Bottom:} Correct partial differential equation along with the identified one obtained by learning $\lambda_1$ and $\lambda_2$.}
\label{fig:Burgers_CT_identification}
\end{figure}

To illustrate the effectiveness of our approach, we have created a training data-set by randomly generating $N = 2,000$ points across the entire spatio-temporal domain from the exact solution corresponding to $\lambda_1 = 1.0$ and $\lambda_2 = 0.01/\pi$. The locations of the training points are illustrated in the top panel of figure \ref{fig:Burgers_CT_identification}. This data-set is then used to train a 9-layer deep neural network with 20 neurons per hidden layer by minimizing the mean square error loss of \eqref{eq:MSE_Burgers_CT_inference} using the L-BFGS optimizer \cite{liu1989limited}. Upon training, the network is calibrated to  predict the entire solution $u(t,x)$, as well as the unknown parameters $\lambda = (\lambda_1, \lambda_2)$ that define the underlying dynamics. A visual assessment of the predictive accuracy of the {\em physics informed neural network} is given in the middle and bottom panels of figure \ref{fig:Burgers_CT_identification}. The network is able to identify the underlying partial differential equation with remarkable accuracy, even in the case where the scattered training data is corrupted with 1\% uncorrelated noise.\\

To further scrutinize the performance of our algorithm, we have performed a systematic study with respect to the total number of training data, the noise corruption levels, and the neural network architecture. The results are summarized in tables \ref{tab:Burgers_CT_identification_1} and \ref{tab:Burgers_CT_identification_2}. The key observation here is that the proposed methodology appears to be very robust with respect to noise levels in the data, and yields a reasonable identification accuracy even for noise corruption up to 10\%. This enhanced robustness seems to greatly outperform competing approaches using Gaussian process regression as previously reported in \cite{raissi2017hidden} as well as approaches relying on sparse regression that require relatively clean data for accurately computing numerical gradients \cite{brunton2016discovering}.

\begin{table}[!t]
\centering
\resizebox{\textwidth}{!}{
\begin{tabular}{|l||cccc||cccc|} \hline
& \multicolumn{4}{c||}{\% error in $\lambda_1$} & \multicolumn{4}{c|} {\% error in $\lambda_2$} \\ \hline
\diagbox{$N_u$}{noise} & 0\% & 1\% & 5\% & 10\% & 0\% & 1\% & 5\% & 10\% \\ \hline\hline
500 &  0.131 & 0.518 & 0.118 & 1.319 & 13.885 & 0.483 & 1.708 & 4.058 \\
1000 & 0.186 & 0.533 & 0.157 & 1.869 & 3.719 & 8.262 & 3.481 & 14.544 \\
1500 & 0.432 & 0.033 & 0.706 & 0.725 & 3.093 & 1.423 & 0.502 & 3.156 \\
2000 & 0.096 & 0.039 & 0.190 & 0.101 & 0.469 & 0.008 & 6.216 & 6.391 \\ \hline
\end{tabular}
}
\caption{{\em Burgers' equation:} Percentage error in the identified parameters $\lambda_1$ and $\lambda_2$ for different number of training data $N$ corrupted by different noise levels. Here, the neural network architecture is kept fixed to 9 layers and 20 neurons per layer.} \label{tab:Burgers_CT_identification_1}
\end{table}

\begin{table}[!t]
\centering
\begin{tabular}{|c||ccc||ccc|} \hline
& \multicolumn{3}{c||}{\% error in $\lambda_1$} & \multicolumn{3}{c|} {\% error in $\lambda_2$} \\ \hline
\diagbox{Layers}{Neurons} & 10 & 20 & 40 & 10 & 20 & 40 \\ \hline\hline
2 & $11.696$ & $2.837$ & $1.679$ & $103.919$ & $67.055$ & $49.186$ \\
4 & $0.332$ & $0.109$ & $0.428$ & $4.721$ & $1.234$ & $6.170$ \\
6 & $0.668$ & $0.629$ & $0.118$ & $3.144$ & $3.123$ & $1.158$ \\
8 & $0.414$ & $0.141$ & $0.266$ & $8.459$ & $1.902$ & $1.552$ \\ \hline
\end{tabular}
\caption{{\em Burgers' equation:} Percentage error in the identified parameters $\lambda_1$ and $\lambda_2$ for different number of hidden layers and neurons per layer. Here, the training data is considered to be noise-free and fixed to $N = 2,000$.} \label{tab:Burgers_CT_identification_2}
\end{table}

\subsubsection{Example (Navier-Stokes Equation)}
Our next example involves a realistic scenario of incompressible fluid flow as described by the ubiquitous Navier-Stokes equations. Navier-Stokes equations describe the physics of many phenomena of scientific and engineering interest. They may be used to model the weather, ocean currents, water flow in a pipe and air flow around a wing. The Navier-Stokes equations in their full and simplified forms help with the design of aircraft and cars, the study of blood flow, the design of power stations, the analysis of the dispersion of pollutants, and many other applications. Let us consider the Navier-Stokes equations in two dimensions\footnote{It is straightforward to generalize the proposed framework to the Navier-Stokes equations in three dimensions (3D).} (2D) given explicitly by
\begin{equation}\label{eq:NavierStokes}
\begin{array}{c}
u_t + \lambda_1 (u u_x + v u_y) = -p_x + \lambda_2(u_{xx} + u_{yy}),\\
v_t + \lambda_1 (u v_x + v v_y) = -p_y + \lambda_2(v_{xx} + v_{yy}),
\end{array}
\end{equation}
where $u(t, x, y)$ denotes the $x$-component of the velocity field, $v(t, x, y)$ the $y$-component, and $p(t, x, y)$ the pressure. Here, $\lambda = (\lambda_1, \lambda_2)$ are the unknown parameters. Solutions to the Navier-Stokes equations are searched in the set of divergence-free functions; i.e.,
\begin{equation}\label{eq:Continuity}
u_x + v_y = 0.
\end{equation}
This extra equation is the continuity equation for incompressible fluids that describes the conservation of mass of the fluid. We make the assumption that
\begin{equation}\label{eq:streamline}
u = \psi_y,\ \ \ v = -\psi_x,
\end{equation}
for some latent function $\psi(t,x,y)$.\footnote{This construction can be generalized to three dimensional problems by employing the notion of vector potentials.} Under this assumption, the continuity equation (\ref{eq:Continuity}) will be automatically satisfied. Given noisy measurements
\[
\{t^i, x^i, y^i, u^i, v^i\}_{i=1}^{N}
\]
of the velocity field, we are interested in learning the parameters $\lambda$ as well as the pressure $p(t,x,y)$. We define $f(t,x,y)$ and $g(t,x,y)$ to be given by
\begin{eqnarray}\label{eq:NavierStokes_RHS}
\begin{array}{c}
f := u_t + \lambda_1 (u u_x + v u_y) + p_x - \lambda_2(u_{xx} + u_{yy}),\\
g := v_t + \lambda_1 (u v_x + v v_y) + p_y - \lambda_2(v_{xx} + v_{yy}),
\end{array}
\end{eqnarray}
and proceed by jointly approximating $\begin{bmatrix}
\psi(t,x,y) & p(t,x,y)
\end{bmatrix}$ using a single neural network with two outputs. This prior assumption along with equations (\ref{eq:streamline}) and (\ref{eq:NavierStokes_RHS}) results into a \emph{physics informed neural network} $\begin{bmatrix}
f(t,x,y) & g(t,x,y)
\end{bmatrix}$. The parameters $\lambda$ of the Navier-Stokes operator as well as the parameters of the neural networks $\begin{bmatrix}
\psi(t,x,y) & p(t,x,y)
\end{bmatrix}$ and $\begin{bmatrix}
f(t,x,y) & g(t,x,y)
\end{bmatrix}$ can be trained by minimizing the mean squared error loss
\begin{eqnarray}
MSE &:=& \frac{1}{N}\sum_{i=1}^{N} \left(|u(t^i,x^i,y^i) - u^i|^2 + |v(t^i,x^i,y^i) - v^i|^2\right) \\
    & +& \frac{1}{N}\sum_{i=1}^{N} \left(|f(t^i,x^i,y^i)|^2 + |g(t^i,x^i,y^i)|^2\right).\nonumber
\end{eqnarray}
Here we consider the prototype problem of incompressible flow past a circular cylinder; a problem known to exhibit rich dynamic behavior and transitions for different regimes of the Reynolds number $Re = u_{\infty}D/\nu$. Assuming a non-dimensional free stream velocity $u_{\infty} = 1$, cylinder diameter $D = 1$, and kinematic viscosity $\nu = 0.01$, the system exhibits a periodic steady state behavior characterized by a asymmetrical vortex shedding pattern in the cylinder wake, known as the K\'arm\'an vortex street \cite{von1963aerodynamics}.\\

\begin{figure}[!t]
\includegraphics[width = 1.0\textwidth]{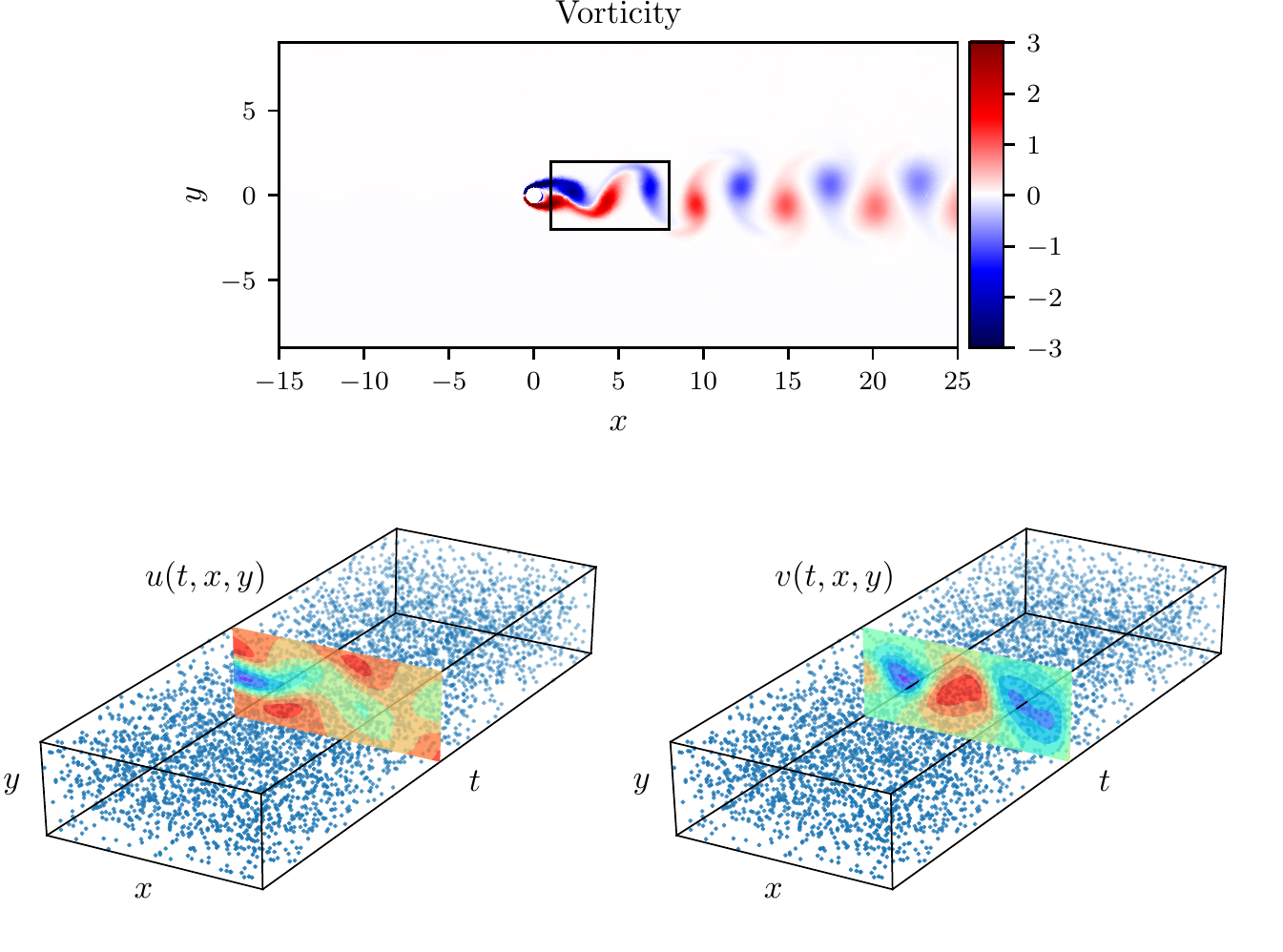}
\caption{{\em Navier-Stokes equation:} {\it Top:} Incompressible flow and dynamic vortex shedding past a circular cylinder at $Re=100$. The spatio-temporal training data correspond to the depicted rectangular region in the cylinder wake. {\it Bottom:} Locations of training data-points for the the stream-wise and transverse velocity components, $u(t,x,y)$ and $v(t,x,t)$, respectively.} 
\label{fig:NavierStokes_data}
\end{figure}

To generate a high-resolution data set for this problem we have employed the spectral/{\em hp}-element solver NekTar \cite{karniadakis2013spectral}. Specifically, the solution domain is discretized in space by a tessellation consisting of 412 triangular elements, and within each element the solution is approximated as a linear combination of a tenth-order hierarchical, semi-orthogonal Jacobi polynomial expansion \cite{karniadakis2013spectral}. We have assumed a uniform free stream velocity profile imposed at the left  boundary, a zero pressure outflow condition imposed at the right boundary located 25 diameters downstream of the cylinder, and periodicity for the top and bottom boundaries of the $[-15,25]\times[-8,8]$ domain. We integrate equation \eqref{eq:NavierStokes} using a third-order stiffly stable scheme \cite{karniadakis2013spectral} until the system reaches a periodic steady state, as depicted in figure \ref{fig:NavierStokes_data}(a). In what follows, a small portion of the resulting data-set corresponding to this steady state solution will be used for model training, while the remaining data will be used to validate our predictions. For simplicity, we have chosen to confine our sampling in a rectangular region downstream of cylinder as shown in figure \ref{fig:NavierStokes_data}(a).\\

Given scattered and potentially noisy data on the stream-wise $u(t,x,y)$ and transverse $v(t,x,y)$ velocity components, our goal is to identify the unknown parameters $\lambda_1$ and $\lambda_2$, as well as to obtain a qualitatively accurate reconstruction of the entire pressure field $p(t,x,y)$ in the cylinder wake, which by definition can only be identified up to a constant. To this end, we have created a training data-set by randomly sub-sampling the full high-resolution data-set. To highlight the ability of our method to learn from scattered and scarce training data, we have chosen $N=5,000$, corresponding to a mere 1\% of the total available data as illustrated in figure \ref{fig:NavierStokes_data}(b). Also plotted are representative snapshots of the predicted velocity components $u(t,x,y)$ and $v(t,x,y)$ after the model was trained. The  neural network architecture used here consists of 9 layers with 20 neurons in each layer.\\

\begin{figure}[!t]
\includegraphics[width = 1.0\textwidth]{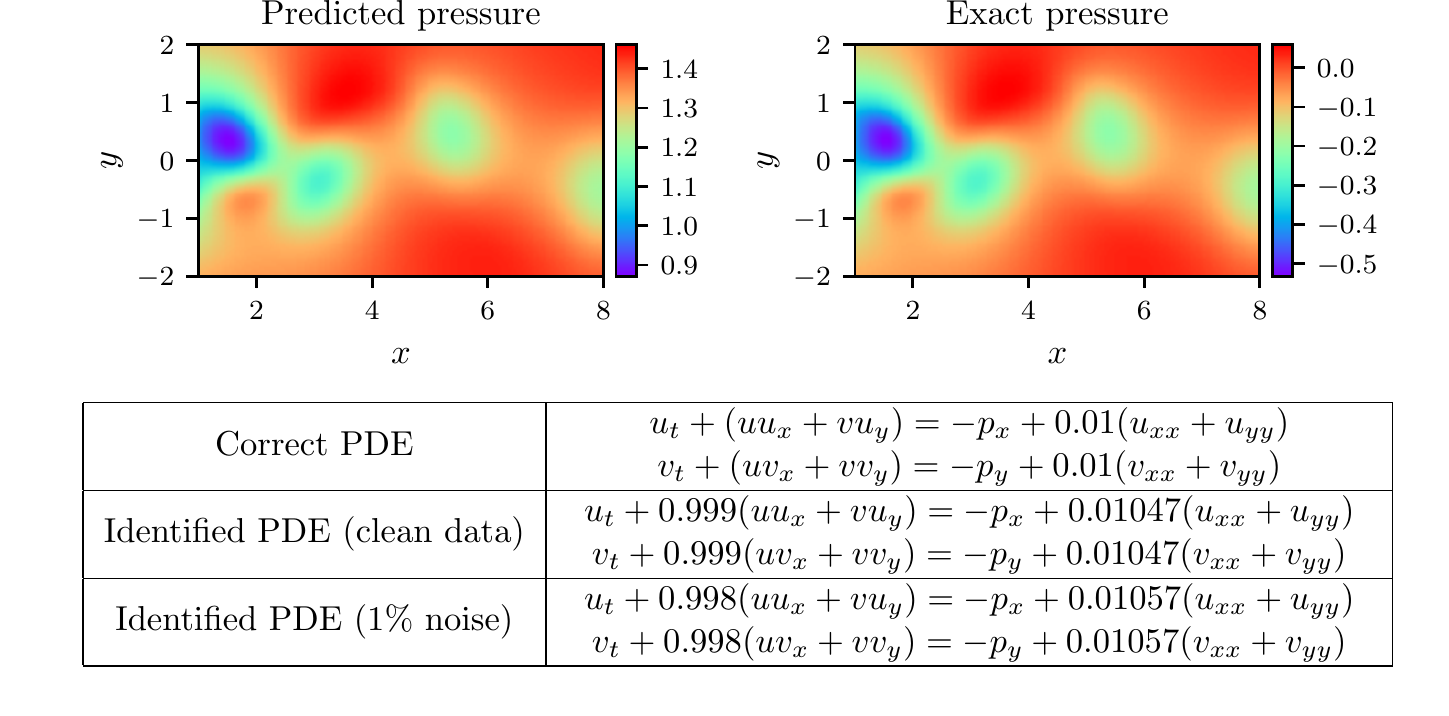}
\caption{{\em Navier-Stokes equation:} {\it Top:} Predicted versus exact instantaneous pressure field $p(t,x,y)$ at a representative time instant. By definition, the pressure can be recovered up to a constant, hence justifying the different magnitude between the two plots. This remarkable qualitative agreement highlights the ability of {\em physics-informed neural networks} to identify the entire pressure field, despite the fact that no data on the pressure are used during model training. {\it Bottom:} Correct partial differential equation along with the identified one obtained by learning $\lambda_1, \lambda_2$ and $p(t,x,y)$.}
\label{fig:NavierStokes_prediction}
\end{figure}

A summary of our results for this example is presented in figure \ref{fig:NavierStokes_prediction}. We observe that the {\em physics informed neural network} is able to correctly identify the unknown parameters $\lambda_1$ and $\lambda_2$ with very high accuracy even when the training data was corrupted with noise. Specifically, for the case of noise-free training data, the error in estimating $\lambda_1$ and $\lambda_2$ is 0.078\%, and 4.67\%, respectively. The predictions remain robust even when the training data are corrupted with 1\% uncorrelated Gaussian noise, returning an error of 0.17\%, and 5.70\%, for $\lambda_1$ and $\lambda_2$, respectively.\\

A more intriguing result stems from the network's ability to provide a qualitatively accurate prediction of the entire pressure field $p(t,x,y)$ in the absence of any training data on the pressure itself. A visual comparison against the exact pressure solution is presented in figure \ref{fig:NavierStokes_prediction} for a representative pressure snapshot. Notice that the difference in magnitude between the exact and the predicted pressure is justified by the very nature of the Navier-Stokes system, as the pressure field is only identifiable up to a constant. This result of inferring a continuous quantity of interest from auxiliary measurements by leveraging the underlying physics is a great example of the enhanced capabilities that {\em physics informed neural networks} have to offer, and highlights their potential in solving high-dimensional inverse problems.\\

Our approach so far assumes availability of scattered data throughout the entire spatio-temporal domain. However, in many cases of practical interest, one may only be able to observe the system at distinct time instants. In the next section, we introduce a different approach that tackles the data-driven discovery problem using only two data snapshots. We will see how, by leveraging the classical Runge-Kutta time-stepping schemes, one can construct discrete time {\em physics informed neural networks} that can retain high predictive accuracy even when the temporal gap between the data snapshots is very large.

\section{Discrete Time Models}
We begin by applying the general form of Runge-Kutta methods with $q$ stages to equation \eqref{eq:PDE} and obtain
\begin{equation}\label{eq:RungeKutta}
\arraycolsep=1.0pt\def\arraystretch{1.5}
\begin{array}{ll}
u^{n+c_i} = u^n - \Delta t \sum_{j=1}^q a_{ij} \mathcal{N}[u^{n+c_j};\lambda], \ \ i=1,\ldots,q,\\
u^{n+1} = u^{n} - \Delta t \sum_{j=1}^q b_j \mathcal{N}[u^{n+c_j};\lambda].
\end{array}
\end{equation}
Here, $u^{n+c_j}(x) = u(t^n + c_j \Delta t, x)$ for $j=1, \ldots, q$. This general form encapsulates both implicit and explicit time-stepping schemes, depending on the choice of the parameters $\{a_{ij},b_j,c_j\}$. Equations (\ref{eq:RungeKutta}) can be equivalently expressed as
\begin{equation}
\arraycolsep=1.0pt\def\arraystretch{1.5}
\begin{array}{ll}
u^{n} = u^n_i, \ \ i=1,\ldots,q,\\
u^{n+1} = u^{n+1}_{i}, \ \ i=1,\ldots,q.
\end{array}
\end{equation}
where
\begin{equation}\label{eq:RungeKutta_identification_rearranged}
\arraycolsep=1.0pt\def\arraystretch{1.5}
\begin{array}{ll}
u^n_i := u^{n+c_i} + \Delta t \sum_{j=1}^q a_{ij} \mathcal{N}[u^{n+c_j};\lambda], \ \ i=1,\ldots,q,\\
u^{n+1}_{i} := u^{n+c_i} + \Delta t \sum_{j=1}^q (a_{ij} - b_j) \mathcal{N}[u^{n+c_j};\lambda], \ \ i=1,\ldots,q.
\end{array}
\end{equation}
We proceed by placing a multi-output neural network prior on
\begin{equation}\label{eq:RungeKutta_PU_prior}
\begin{bmatrix}
u^{n+c_1}(x), \ldots, u^{n+c_q}(x)
\end{bmatrix}.
\end{equation}
This prior assumption along with equations (\ref{eq:RungeKutta_identification_rearranged}) result in two \emph{physics informed neural networks}
\begin{equation}\label{eq:RungeKutta_PI_prior_1}
\begin{bmatrix}
u^{n}_1(x), \ldots, u^{n}_q(x), u^{n}_{q+1}(x)
\end{bmatrix},
\end{equation}
and
\begin{equation}\label{eq:RungeKutta_PI_prior_2}
\begin{bmatrix}
u^{n+1}_1(x), \ldots, u^{n+1}_q(x), u^{n+1}_{q+1}(x)
\end{bmatrix}.
\end{equation}
Given noisy measurements at two distinct temporal snapshots $\{\bm{x}^{n}, \bm{u}^{n}\}$ and $\{\bm{x}^{n+1}, \bm{u}^{n+1}\}$ of the system at times $t^{n}$ and $t^{n+1}$, respectively, the shared parameters of the neural networks \eqref{eq:RungeKutta_PU_prior}, \eqref{eq:RungeKutta_PI_prior_1}, and \eqref{eq:RungeKutta_PI_prior_2} along with the parameters $\lambda$ of the differential operator can be trained by minimizing the sum of squared errors
\begin{equation}\label{eq:SSE_DT_identification}
SSE = SSE_n + SSE_{n+1},
\end{equation}
where
\[
SSE_n := \sum_{j=1}^q \sum_{i=1}^{N_n} |u^n_j(x^{n,i}) - u^{n,i}|^2,
\]
and
\[
SSE_{n+1} := \sum_{j=1}^q \sum_{i=1}^{N_{n+1}} |u^{n+1}_j(x^{n+1,i}) - u^{n+1,i}|^2.
\]
Here, $\bm{x}^n = \left\{x^{n,i}\right\}_{i=1}^{N_n}$, $\bm{u}^n = \left\{u^{n,i}\right\}_{i=1}^{N_n}$, $\bm{x}^{n+1} = \left\{x^{n+1,i}\right\}_{i=1}^{N_{n+1}}$, and $\bm{u}^{n+1} = \left\{u^{n+1,i}\right\}_{i=1}^{N_{n+1}}$.

\subsection{Example (Burgers' Equation)}
\begin{figure}[!t]
\includegraphics[width = 1.0\textwidth]{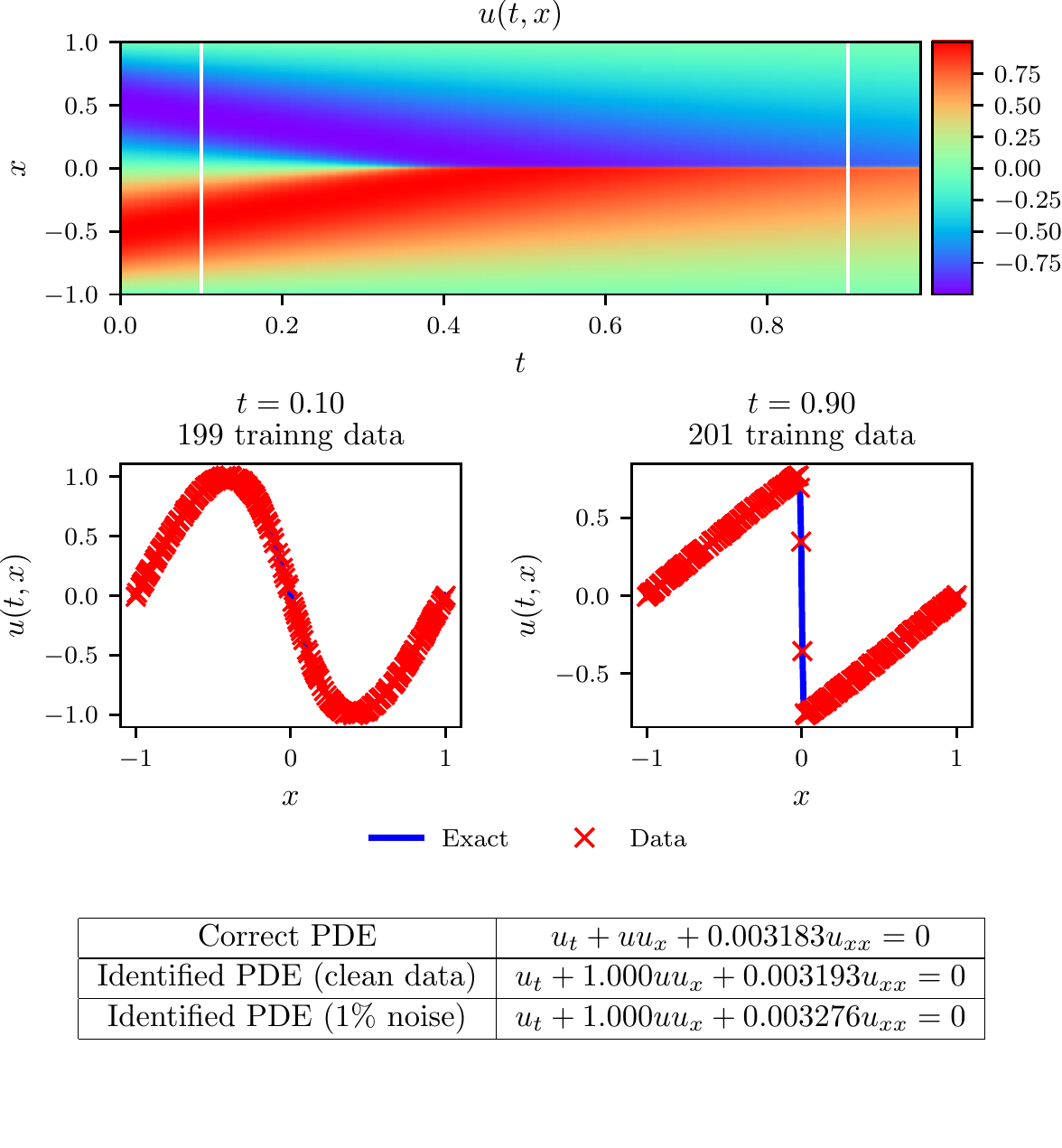}
\vspace*{-1.5cm}
\caption{{\em Burgers equation:} {\it Top:} Solution $u(t,x)$ along with the temporal locations of the two training snapshots. {\it Middle:} Training data and exact solution corresponding to  the two temporal snapshots depicted by the dashed vertical lines in the top panel. {\it Bottom:} Correct partial differential equation along with the identified one obtained by learning $\lambda_1, \lambda_2$.}
\label{fig:Burgers_DT_identification}
\end{figure}

Let us illustrate the key features of this method through the lens of the Burgers' equation. Recall the equation's form
\begin{equation}\label{eq:Burgers_DT_identification}
u_t + \lambda_1 u u_x - \lambda_2 u_{xx} = 0,
\end{equation}
and notice that the nonlinear operator in equation \eqref{eq:RungeKutta_identification_rearranged} is given by
\[
\mathcal{N}[u^{n+c_j}] = \lambda_1 u^{n+c_j} u^{n+c_j}_x - \lambda_2 u^{n+c_j}_{xx}.
\]
Given merely two training data snapshots, the shared parameters of the neural networks \eqref{eq:RungeKutta_PU_prior}, \eqref{eq:RungeKutta_PI_prior_1}, and \eqref{eq:RungeKutta_PI_prior_2} along with the parameters $\lambda = (\lambda_1, \lambda_2)$ of the Burgers' equation can be learned by minimizing the sum of squared errors in equation \eqref{eq:SSE_DT_identification}. Here, we have created a training data-set comprising of $N_n=199$ and $N_{n+1}=201$ spatial points by randomly sampling the exact solution at time instants $t^n=0.1$ and $t^{n+1}=0.9$, respectively. The training data are shown in the top and middle panel of figure \ref{fig:Burgers_DT_identification}. The neural network architecture used here consists of 4 hidden layers with 50 neurons each, while the number of Runge-Kutta stages is empirically chosen to yield a temporal error accumulation of the order of machine precision $\epsilon$ by setting\footnote{This is motivated by the theoretical error estimates for implicit Runge-Kutta schemes suggesting a truncation error of $\mathcal{O}(\Delta{t}^{2q})$ \cite{iserles2009first}.}
\begin{equation}\label{eq:Runge-Kutta_stages}
q = 0.5\log{\epsilon}/\log(\Delta{t}),
\end{equation}
where the time-step for this example is $\Delta{t}=0.8$. The bottom panel of figure \ref{fig:Burgers_DT_identification} summarizes the identified parameters $\lambda = (\lambda_1, \lambda_2)$ for the cases of noise-free data, as well as noisy data with 1\% of Gaussian uncorrelated noise corruption. For both cases, the proposed algorithm is able to learn the correct parameter values $\lambda_1=1.0$ and  $\lambda_2=0.01/\pi$ with remarkable accuracy, despite the fact that the two data snapshots used for training are very far apart, and potentially describe different regimes of the underlying dynamics.\\

A sensitivity analysis is performed to quantify the accuracy of our predictions with respect to the gap between the training snapshots $\Delta{t}$, the noise levels in the training data, and the \emph{physics informed neural network} architecture. As shown in table \ref{tab:Burgers_DT_identification_1}, the proposed algorithm is quite robust to both $\Delta{t}$ and the noise corruption levels, and it consistently returns reasonable estimates for the unknown parameters. This robustness is mainly attributed to the flexibility of the underlying implicit Runge-Kutta scheme to admit an arbitrarily high number of stages, allowing the data snapshots to be very far apart in time, while not compromising the accuracy with which the nonlinear dynamics of equation \eqref{eq:Burgers_DT_identification} are resolved. This is the key highlight of our discrete time formulation for identification problems, setting it apart from competing approaches \cite{raissi2017hidden,brunton2016discovering}. Lastly, table \ref{tab:Burgers_DT_identification_2} presents the percentage error in the identified parameters, demonstrating the robustness of our estimates with respect to the underlying neural network architecture.

\begin{table}[!t]
\centering
\resizebox{\textwidth}{!}{
\begin{tabular}{|l||cccc||cccc|} \hline
& \multicolumn{4}{c||}{\% error in $\lambda_1$} & \multicolumn{4}{c|} {\% error in $\lambda_2$} \\ \hline
\diagbox{$\Delta{t}$}{noise} & 0\% & 1\% & 5\% & 10\% & 0\% & 1\% & 5\% & 10\% \\ \hline\hline
0.2 & $0.002$ & $0.435$ & $6.073$ & $3.273$ & $0.151$ & $4.982$ & $59.314$ & $83.969$ \\
0.4 & $0.001$ & $0.119$ & $1.679$ & $2.985$ & $0.088$ & $2.816$ & $8.396$ & $8.377$ \\
0.6 & $0.002$ & $0.064$ & $2.096$ & $1.383$ & $0.090$ & $0.068$ & $3.493$ & $24.321$ \\
0.8 & $0.010$ & $0.221$ & $0.097$ & $1.233$ & $1.918$ & $3.215$ & $13.479$ & $1.621$ \\ \hline
\end{tabular}
}
\caption{{\em Burgers' equation:} Percentage error in the identified parameters $\lambda_1$ and $\lambda_2$ for different gap size $\Delta{t}$ between two different snapshots and for different noise levels.} \label{tab:Burgers_DT_identification_1}
\end{table}

\begin{table}[!t]
\centering
\resizebox{\textwidth}{!}{
\begin{tabular}{|c||ccc||ccc|} \hline
& \multicolumn{3}{c||}{\% error in $\lambda_1$} & \multicolumn{3}{c|} {\% error in $\lambda_2$} \\ \hline
\diagbox{Layers}{Neurons} & 10 & 25 & 50 & 10 & 25 & 50 \\ \hline\hline
1 & $1.868$ & $4.868$ & $1.960$ & $180.373$ & $237.463$ & $123.539$ \\
2 & $0.443$ & $0.037$ & $0.015$ & $29.474$ & $2.676$ & $1.561$ \\
3 & $0.123$ & $0.012$ & $0.004$ & $7.991$ & $1.906$ & $0.586$ \\
4 & $0.012$ & $0.020$ & $0.011$ & $1.125$ & $4.448$ & $2.014$ \\ \hline
\end{tabular}
}
\caption{{\em Burgers' equation:} Percentage error in the identified parameters $\lambda_1$ and $\lambda_2$ for different number of hidden layers and neurons in each layer.} \label{tab:Burgers_DT_identification_2}
\end{table}

\subsubsection{Example (Korteweg–de Vries Equation)}
Our final example aims to highlight the ability of the proposed framework to handle governing partial differential equations involving higher order derivatives. Here, we consider a mathematical model of waves on shallow water surfaces; the Korteweg-de Vries (KdV) equation. This equation can also be viewed as Burgers' equation with an added dispersive term. The KdV equation has several connections to physical problems. It describes the evolution of long one-dimensional waves in many physical settings. Such physical settings include shallow-water waves with weakly non-linear restoring forces, long internal waves in a density-stratified ocean, ion acoustic waves in a plasma, and acoustic waves on a crystal lattice. Moreover, the KdV equation is the governing equation of the string in the Fermi-Pasta-Ulam problem \cite{dauxois2008fermi} in the continuum limit. The KdV equation reads as
\begin{equation}\label{eq:KdV}
u_t + \lambda_1 u u_x + \lambda_2 u_{xxx} = 0,
\end{equation}
with $(\lambda_1, \lambda_2)$ being the unknown parameters. For the KdV equation, the nonlinear operator in equations \eqref{eq:RungeKutta_identification_rearranged} is given by
\[
\mathcal{N}[u^{n+c_j}] = \lambda_1 u^{n+c_j} u^{n+c_j}_x - \lambda_2 u^{n+c_j}_{xxx}
\]
and the shared parameters of the neural networks \eqref{eq:RungeKutta_PU_prior}, \eqref{eq:RungeKutta_PI_prior_1}, and \eqref{eq:RungeKutta_PI_prior_2} along with the parameters $\lambda = (\lambda_1, \lambda_2)$ of the KdV equation can be learned by minimizing the sum of squared errors \eqref{eq:SSE_DT_identification}.\\

To obtain a set of training and test data we simulated \eqref{eq:KdV} using conventional spectral methods.
Specifically, starting from an initial condition $u(0,x) = \cos(\pi x)$ and assuming periodic boundary conditions, we have integrated equation \eqref{eq:KdV} up to a final time $t=1.0$ using the Chebfun package \cite{driscoll2014chebfun} with a spectral Fourier discretization with 512 modes and a fourth-order explicit Runge-Kutta temporal integrator with time-step $\Delta{t} = 10^{-6}$. Using this data-set, we then extract two solution snapshots at time $t^n=0.2$ and $t^{n+1}=0.8$, and randomly sub-sample them using $N_n = 199$ and $N_{n+1} = 201$ to generate a training data-set. We then use these data to train a discrete time {\em physics informed neural network} by minimizing the sum of squared error loss of equation \eqref{eq:SSE_DT_identification} using L-BFGS \cite{liu1989limited}. The network architecture used here comprises of 4 hidden layers, 50 neurons per layer, and an output layer predicting the solution at the $q$ Runge-Kutta stages, i.e., $u^{n+c_j}(x)$, $j=1,\dots,q$, where $q$ is computed using equation \eqref{eq:Runge-Kutta_stages} by setting $\Delta{t}=0.6$.\\

\begin{figure}[!t]
\includegraphics[width = 1.0\textwidth]{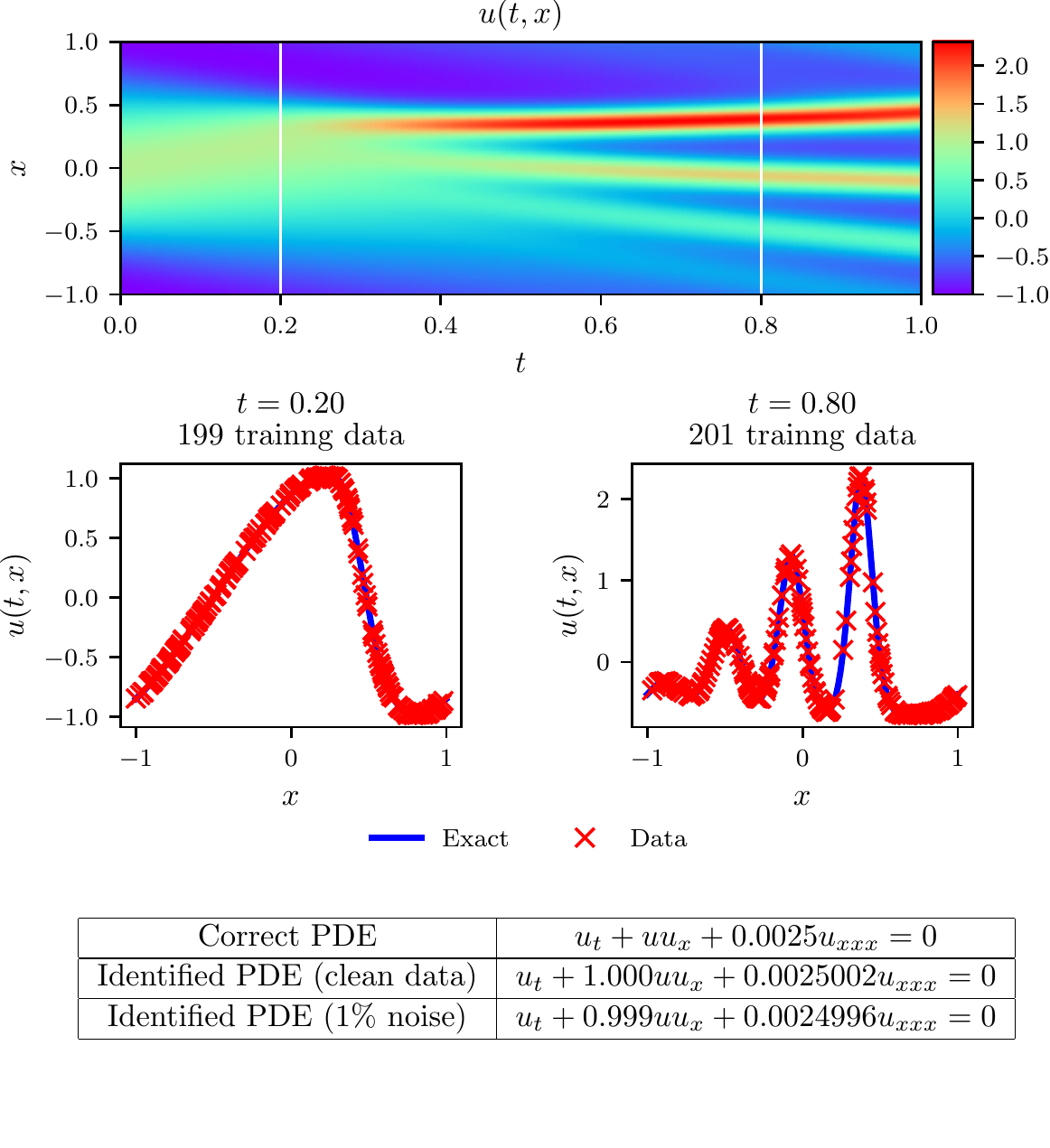}
\vspace*{-1.8cm}
\caption{{\em KdV equation:} {\it Top:} Solution $u(t,x)$ along with the temporal locations of the two training snapshots. {\it Middle:} Training data and exact solution corresponding to  the two temporal snapshots depicted by the dashed vertical lines in the top panel. {\it Bottom:} Correct partial differential equation along with the identified one obtained by learning $\lambda_1, \lambda_2$.}
\label{fig:KdV_DT_identification}
\end{figure}

The results of this experiment are summarized in figure \ref{fig:KdV_DT_identification}. In the top panel, we present the exact solution $u(t,x)$, along with the locations of the two data snapshots used for training. A more detailed overview of the exact solution and the training data is given in the middle panel. It is worth noticing how the complex nonlinear dynamics of equation \eqref{eq:KdV} causes dramatic differences in the form of the solution between the two reported snapshots. Despite these differences, and the large temporal gap between the two training snapshots, our method is able to correctly identify the unknown parameters regardless of whether the training data is corrupted with noise or not. 
Specifically, for the case of noise-free training data, the error in estimating $\lambda_1$ and $\lambda_2$ is 0.023\%, and 0.006\%, respectively, while the case with 1\% noise in the training data returns errors of 0.057\%, and 0.017\%, respectively. 

\section{Summary and Discussion} 

We have introduced {\em physics informed neural networks}, a new class of universal function approximators that is capable of encoding any underlying physical laws that govern a given data-set, and can be described by partial differential equations. In this work, we design data-driven algorithms for discovering dynamic models described by parametrized nonlinear partial differential equations. The inferred models allow us to construct computationally efficient and fully differentiable surrogates that can be subsequently used for different applications including predictive forecasting, control, and optimization. \\

Although a series of promising results was presented, the reader may perhaps agree that this two-part treatise creates more questions than it answers. In a broader context, and along the way of seeking further understanding of such tools, we believe that this work advocates a fruitful synergy between machine learning and classical computational physics that has the potential to enrich both fields and lead to high-impact developments.

\section*{Acknowledgements}
This work received support by the DARPA EQUiPS grant N66001-15-2-4055, the MURI/ARO grant W911NF-15-1-0562, and the AFOSR grant FA9550-17-1-0013. All data and codes used in this manuscript are publicly available on GitHub at \url{https://github.com/maziarraissi/PINNs}.





\bibliographystyle{model1-num-names}
\bibliography{sample.bib}







\end{document}